# Learning-based Stage Verification System in Manual Assembly Scenarios

Xingjian Zhang[1[0009-0003-8137-8098]], Yutong Duan[1[0009-0006-0076-8601]],

Zaishu Chen[1[0009-0003-5054-8736]]

[1] College of Design and Engineering, National University of Singapore, 119077, Singapore
xingjian@u.nus.edu

**Abstract.** In the context of Industry 4.0, effective monitoring of multiple targets and states during assembly processes is crucial, particularly when constrained to using only visual sensors. Traditional methods often rely on either multiple sensor types or complex hardware setups to achieve high accuracy in monitoring, which can be cost-prohibitive and difficult to implement in dynamic industrial environments. This study presents a novel approach that leverages multiple machine learning models to achieve precise monitoring under the limitation of using a minimal number of visual sensors. By integrating state information from identical timestamps, our method detects and confirms the current stage of the assembly process with an average accuracy exceeding 92%. Furthermore, our approach surpasses conventional methods by offering enhanced error detection and visualization capabilities, providing real-time, actionable guidance to operators. This not only improves the accuracy and efficiency of assembly monitoring but also reduces dependency on expensive hardware solutions, making it a more practical choice for modern industrial applications.

**Keywords:** Manual Assembly, Action Recognition, Machine Learning.

## 1 Introduction

With the gradual popularization of related concepts in Industry 4.0, the demand for highly automated production lines in the industry continues to increase, and there are also many manual assembly processes that cannot be replaced in a short period of time [1]. The characteristic of manual assembly process is its high adaptability to product and part conditions, operator skills, and constantly changing production requirements [2]. Assembly verification and quality assurance (QA) rely on people and the inspection and verification of human quality assurance can be time-consuming, which often cannot guarantee the high efficiency of the production line in a stable manner [3].

However, human workers are not only the main subject of this task, they are also the most dynamic factor in any advanced intelligent manufacturing system [4]. Based on such human-centered mindset, the main task is to understand human behavior that leads to achieving the optimal work performance. Instead of utilizing robotic arm to transform the original assembly process into a human-machine collaborative process, this field of research is especially significant for absolute manual manufacturing plants, where most production tasks are carried out by several workers in a production line that follows complex work routines.

There exist many studies today that address the challenges and need of quality assurance systems on the production assembly line. The deployment of cameras to capture the workflow on the assembly line and the use of computer vision, machine learning and deep learning methods to provide insights is one of the common solutions that exits [5]. Some of the reasons for using vision-based methods is the ability to monitor the assembly in real-time through live image capturing thus reducing the burden of physical human inspection.

In the meantime, Manual assembly processes exhibit high adaptability to product and parts conditions, operators' skills, and changing production requirements. Traditional assembly verification and quality assurance heavily rely on human inspectors, leading to time-consuming processes and suboptimal results[6]. Meanwhile, the reality is that even after receiving industrial product assembly training for factory employees, assembly sequence errors may still occur [7]. At this point, a set of equipment is needed to monitor the actions of factory employees in real time and conduct assembly quality inspections.

In order to achieve error detection and process recording in the future, the first step is to achieve accurate detection of the actions of workers and assembly items during the assembly process [8]. In some specific scenarios, it is also necessary to know the angles of some objects to ensure that they are installed in the correct way. In each stage of assembly, the main focus is on the hand movements of workers, with a relatively fixed range of activities. However, there may be some details that are difficult to define simply due to different operating habits. Therefore, a deep learning approach is chosen to classify them, ultimately achieving full process action monitoring and error detection [9]. At the same time, in order to make the system suitable for workers or supervision, a unified and complete upper computer monitoring system has been developed based on Qt, which can achieve timely feedback of information and generation of operation reports [10].

Based on such application scenarios, a visual assistance system that can provide clear guidance and assist workers in assembly is one of the excellent solutions. The research results of this project have important practical significance and application value, and need to simultaneously apply professional knowledge such as computer vision, deep learning, computer networks, and control system stability. The core sensor only requires a camera with depth data, which is easy to deploy and has a more convenient feature compared to other assembly assistance solutions. It has a good application market and prospects.

## 2  Related Works

### 2.1  Theoretical Framework

How to better integrate manual assembly tasks with artificial intelligence has been a hot topic in the industry in recent years. Throughout the entire work process, the focus is on not treating workers as robots, but still making them the core of assembly production lines [11]. In recent years, domestic and foreign research has mainly focused on the following aspects: 1) the application of artificial intelligence in motion tracking, 2) the focus of artificial intelligence in manual assembly tasks, and 3) the focus of AI algorithms for each task (mostly CNN) [12]. Prinz and Fahle et al. focused on the adaptability of workers with different levels of proficiency to the system, quantifying the operator's level by specifying a series of indicators. We will consider this in subsequent system design but not as a focus.

In terms of action detection, Zamora et al. applied subject working characteristic curves to graphically validate the performance of the classification model from two aspects: frame performance and sequence performance, and verified the accuracy of machine learning methods for action detection [13]. Meanwhile, the action detection process is a repetitive process of prediction and verification[14], ensuring high real-time recognition of the algorithm through high-speed sampling.

In terms of target recognition, the YOLOv5 model developed by ultralytics has good accuracy and has been widely used. YOLOv5 can achieve a detection confidence level of over 92% through image restoration of real tools and components, as well as enhanced data on real images [15] [16]. Due to the fact that the recognition targets of this assembly assistance system vary completely depending on the assembly object, a certain number of self- labeled datasets can achieve good recognition results.

## 2.2    Current Status of Common Assembly Verification System

The robotic arm is widely used in modern production lines for assembly tasks. It offers advantages such as speed, accuracy, consistency, and a 22% increase in efficiency when working alongside humans [17]. Additionally, robotic arms can operate continuously, reducing labor costs. However, drawbacks include complex control schemes, time-consuming parameter tuning, and the need for specialized worker training, which contradicts the "people-centered" approach [18].

Projector-based interaction has also been a common solution to improve assembly efficiency. It provides intuitive assembly guidance, real-time feedback, and error correction, helping operators identify and resolve issues quickly, thus reducing errors and rework [19]. However, projectors are costly, require specific lighting conditions, and are often constrained by the working area [20].

Current assembly assistance systems often use Kinect to track skeletal joint points, training convolutional neural networks to recognize assembly actions [21]. While bone data is effective for coarse movements like arm gestures, finer finger movements benefit more from data collected via EMG and IMU sensors [22]. Combining multiple sensor data and applying oversampling to address imbalanced datasets further enhances system reliability [23].

Additionally, as open-source models lack angle detection capabilities, Jack et al.

proposed using special markers and stereo angles for real-time calculations. However, this method suffers from low stability and accuracy [24].

## 2.3  Object Detection

In terms of the definition of object detection, it is a computer vision task aimed at locating objects in a given image and identifying the category of each object. Now, object detection is divided into two series - RCNN series and YOLO series. YOLO analyzes object detection as a regression problem, completing direct input from the original image to object position and category output. It achieves high real-time performance and is more suitable for this project [25].

People have been working on detecting objects in a more traditional way for a long time. Traditional computer vision methods typically rely on handcrafted features to represent objects in images, such as edges, corners, and textures. These features are then used for object detection by matching them with known templates or classifiers. Common feature descriptors include SIFT, SURF, and HOG. Object detection is often achieved through classifiers like Support Vector Machines (SVMs) or AdaBoost, which are trained to recognize objects using handcrafted features. Object detectors, such as the Viola-Jones detector, scan images using sliding windows and cascaded classifiers to identify potential objects [26].

In contrast, modern deep learning methods, particularly Convolutional Neural Networks (CNNs), have gained popularity for object detection tasks. CNNs can automatically learn feature representations from data and output object positions and categories. Popular object detection networks include the RCNN series (e.g., Faster R-CNN, Mask R-CNN), YOLO series (e.g., YOLOv3, YOLOv4), and SSD (Single Shot MultiBox Detector). These networks typically use CNNs as feature extractors and build detection heads on top to output object positions and categories.

## 2.4  Gesture Recognition

In this project, another technology that holds almost equal importance to object detection is action recognition, particularly that of hand gestures. This requires first identifying the positions of the hands and standardizing their coordinates, followed by action recognition based on these coordinates.

However, in real-world manual assembly scenarios, hands may undergo various poses and deformations during actions, while it's difficult to avoid occlusion caused by other objects or the hands themselves, which may lead to loss of information or incompleteness, thus increasing the complexity of recognition. Moreover, in the specific deployment process, there are even more complex issues to be addressed. Effective training and optimization strategies are crucial for completing this task [27].

### 2.5 Angle Detection

Accurate angle detection during assembly processes is critical for ensuring product quality and assembly precision, particularly in industries such as automotive, electronics, and aerospace [28]. Many studies have focused on the development of methods and technologies that enable real-time and precise measurement of angles in various manufacturing tasks. These technologies are integral to both manual and automated assembly processes, helping reduce errors, improve efficiency, and ensure the consistency of assembled products.

Machine learning models can identify patterns in the assembly process, including angular deviations missed by traditional methods. Wang et al. (2021) used CNNs to monitor handheld tool angles via real-time video from a flexible production line, achieving high accuracy in detecting subtle angle changes, making it ideal for dynamic assembly environments [29].

## 3 Method

### 3.1 Research Framework

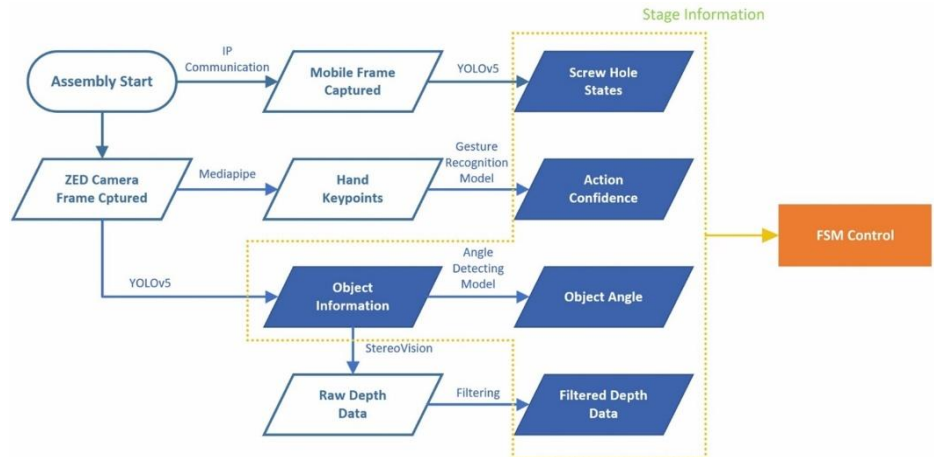

**Fig 1.** Overview of the learning-based stage verification system for manual assembly. It captures frames with ZED and mobile cameras, detects hand keypoints via Mediapipe and object positions with YOLOv5, and outputs screw hole states, action confidence, object angle, and filtered depth data. These are then fed into a Finite State Machine to guide the assembly process.

The whole function of the vision-based assembly stage verification system proposed in this project consists of two components: hand tracking and object detection. Hand tracking and action recognition are used to monitor whether the worker is performing the correct assembly action at that stage and whether the assembly action at that stage has been completed, while object detection is used to monitor whether the worker is performing the correct assembly sequence of the parts and whether the parts are placed in the correct position.

As shown in Fig.1, various detection schemes are used to gather information for assembly. Mediapipe detects hand keypoints, followed by an LSTM model predicting action confidence. YOLOv5 results are used to build a CNN-based angle detection model. Simultaneously, raw depth data from StereoVision is filtered and incorporated for stage verification.

### 3.2 Angle Detection Model

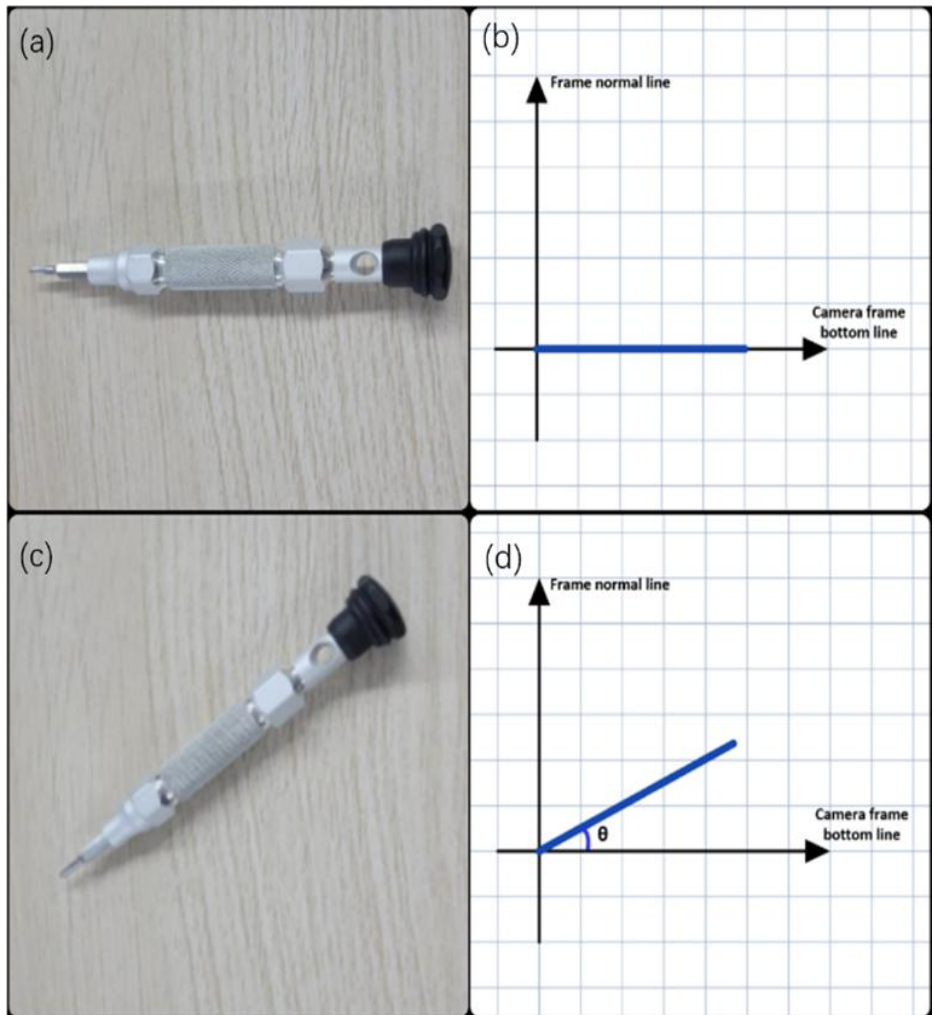

**Fig 2. Normalization of detection object angle.** By defining the object at (a) as the reference angle(b) for its pose, the relative angle of the object in pose (c) is (d).

In YOLO detection, we can only obtain the positional information and confidence score of an object. However, in practical application scenarios, if the angle and position of certain components are improperly assembled, it may affect the assembly of other components. Currently, there are some works on object angle detection, such as applications in the field of natural object detection [30], in palm detection [31]. In the field of part assembly, there is a need to detect specific items; however, the model training in research of Z. Wang et al., it requires a large annotated dataset [32]. Therefore, this project develops a self-supervised angle detection method based on convolutional neural networks, which can be easily trained on a specific set of items and is highly suitable for object angle detection in the assembly field as shown in Fig. 2.

In summary, a convolutional neural network was built for pose detection in the project, and self-supervised learning combined with PyTorch and OpenCV was used to train a convolutional neural network that can recognize specific object placement angles. The specific steps are as follows: Firstly, it is necessary to define the placement state of the recognized object at 0 degrees (which will serve as the basis for subsequent angle determination). Then, use OpenCV to randomly rotate the image by a certain angle, define this angle as x, and use the rotated image as input for the convolutional neural network. The output is the fitted value of the rotation angle, expressed as y. Calculate the absolute value of the loss as x-y, perform gradient descent, and then perform backpropagation.

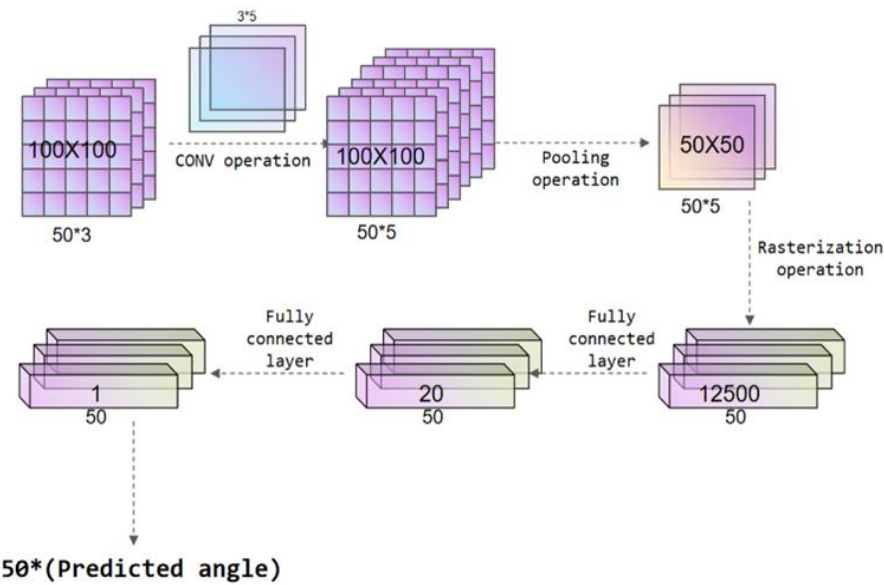

**Fig 3. Angle detection network structure.** Feature extraction is performed on the image using convolutional layers to obtain object edge information, followed by merging and regression of features through fully connected layers.

When using this CNN model (as shown in Fig.3), it is necessary to obtain the region where the object is detected before angle detection through YOLO. Use the resize function of OpenCV to adjust the image size and convert it to the standard input format of a neural network. The output of the model will be the detected angle.

### 3.3 Object Detection

In this project, YOLOv5 was employed for object detection of specific parts using the ZED 2i camera. A total of 1088 training images were prepared from different angles and under varying lighting conditions. The parts in all images were annotated using the Roboflow platform, generating annotation files containing the names and coordinates of two labels in each image. Each image in the dataset was preprocessed to create corresponding annotation files, which include the index of the object category, the normalized coordinates of the target box's center point, and the normalized width and height of the target box.

Roboflow's Health Check function was utilized to analyze the sample distribution across different categories, and adjustments were made to ensure that the number of samples in each category reached the optimal ratio. The dataset was then automatically partitioned into training, validation, and test sets, with respective ratios of 73%, 19%, and 9%. Following this, Roboflow's data augmentation function was applied to expand the dataset quantitatively. As a result, the dataset used for training consisted of a total of 2666 images.

Given that larger input sizes typically lead to improved detection accuracy but also longer detection times, we opted for an input size of $416 \times 416$ pixels to strike a balance between speed and accuracy—both of which are critical for the monitoring system. Data augmentation was applied to both models by randomly altering the saturation, brightness, and hue of the training images, further increasing the diversity of the dataset.

While we initially considered using YOLOv8 models due to their potentially superior performance, the computational resources available in the lab were quite limited [33]. As a result, we opted for YOLOv5x, which offered a balance between performance and trainability within our resource constraints. After fine-tuning the parameters, the YOLOv5x model achieved a mean average precision (mAP) of 92.45%, with each category achieving mAP of at least 90.32%. This performance metric indicated a satisfactory level of detection accuracy for all object categories, making YOLOv5x suitable for our object detection needs.

The testing results confirmed that the fine-tuned YOLOv5x models exhibited high detection accuracy, a critical factor for their subsequent integration into the angle detection and object recognition stages. Despite resource limitations, the results demonstrated that the YOLOv5x model met the project's accuracy requirements, negating the need for further model adjustments or the use of more resource-intensive models like YOLOv8.

### 3.4 Distributed Solution for Screw Detection

In the detection process described above, accurate and stable detection of various components in mechanical hard drives has been achieved. However, improving the detection of screws through model configuration adjustments remains challenging. The fundamental reason lies in the fixed focal length of the depth camera we use. While other items can be clearly detected at the normal operating distance, screws and screw holes are too small, making them difficult to annotate or identify reliably.

A practical solution to this issue is to find a camera with a suitable focal length or one capable of dynamically adjusting its focal length. This would allow it to operate at closer distances to the target area, enabling the clear detection of screws and screw holes. The detected information can then be transmitted back to the main control unit, facilitating precise decisions during the assembly stage.

Given specific application scenarios, the most suitable approach is to use a smartphone camera. Smartphones can dynamically adjust the focal length and perform close-range detection of screws and screw holes effectively.

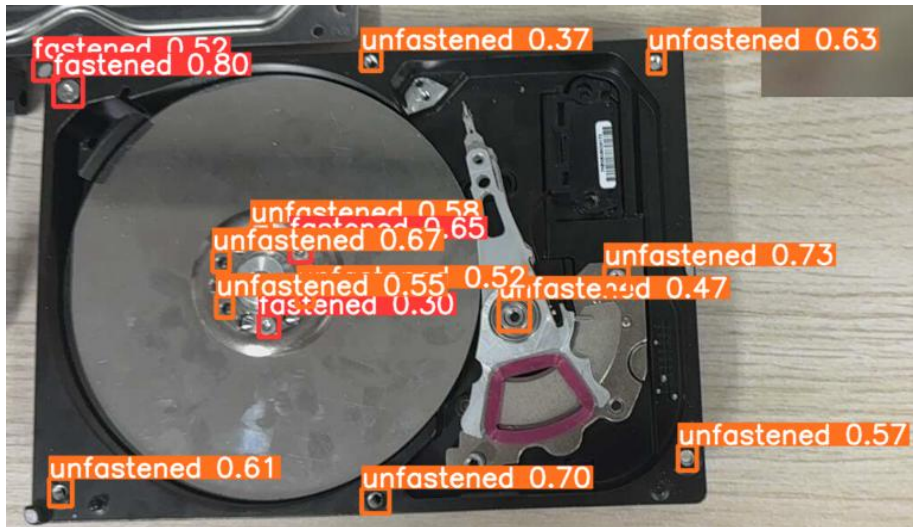

**Fig 4. Screw condition detection result.** The one-shot detection has shown sensitive awareness of all potential objects with relatively low confidence.

The YOLO model's detection logic is based on convolutional neural networks (CNNs), where multiple convolution and pooling operations are applied to the input images, producing feature maps used to predict object locations on these maps. This same logic can be applied to screw detection. In our case, we aim to detect the state of screw holes. The goal is to use a neural network that can identify whether a screw hole is empty, in the process of assembly, or fully assembled with a screw. To achieve this, we collected images of screw holes under these three conditions and annotated them accordingly.

A one-shot detection result is shown in Fig 4, which displays the detection of screw hole states. As seen in the image, nearly all screw holes have been successfully detected; however, the confidence levels are generally lower. This outcome can be attributed to several factors, such as the relatively small size of the dataset and the difficulty of consistently annotating small objects like screws and screw holes. Additionally, the model's ability to precisely identify such fine details may be limited by the camera's fixed focal length and resolution [34].

Nonetheless, despite these challenges, the model demonstrates strong robustness against interference, especially in scenarios where workers may inadvertently attempt to tighten screws without actually picking them up as demonstrated in Fig. 5. In such cases, the model shows a low false positive rate, effectively distinguishing between actual and mistaken actions.

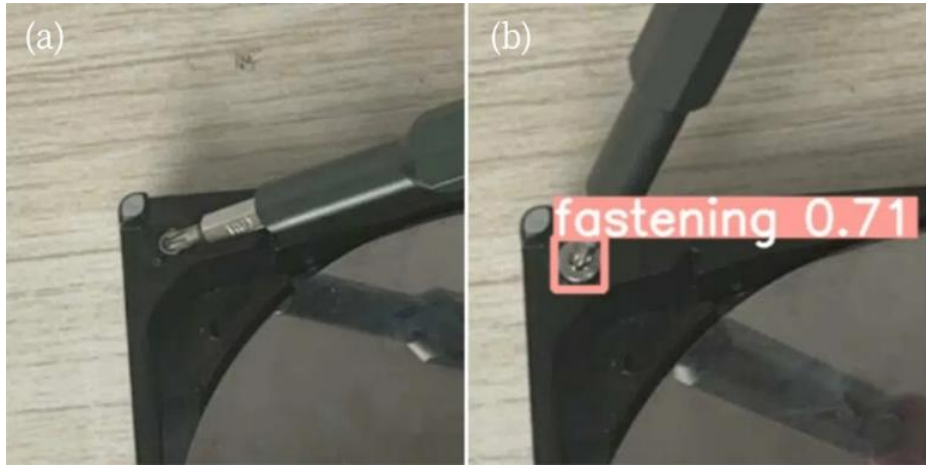

**Fig 5. Comparison of model performance under cheating action.** A worker tried to fasten with no screw, but the model didn't recognize it as fastening.

### 3.5 Gesture Recognition

In addition to object detection, action recognition—particularly hand gesture recognition—plays a critical role in this project. The process begins with identifying the positions of the hands and standardizing their coordinates, followed by action recognition based on these key points.

In real-world manual assembly scenarios, hands often assume a wide variety of poses and deformations during actions, and occlusion by objects or the hands themselves is common. This leads to information loss or incomplete data, further complicating the recognition process [35]. Additionally, more complex issues arise during the deployment process, making effective training and optimization strategies essential to achieving reliable results.

By leveraging the Mediapipe API, we can accurately track and provide real-time feedback on 20 key points for each hand. Additionally, by utilizing the holistic module, key points for the torso are also annotated, enhancing the portability of the system and facilitating potential applications in a broader range of assembly scenarios.

Once hand key points are tracked and their coordinates are acquired, further gesture recognition is necessary. In different assembly tasks, workers exhibit distinct hand gestures, requiring specialized detection schemes for different actions. For this project, a mechanical hard drive was selected as the assembly object. One advantage of this choice is the significant shape variation between its components, making object detection straightforward. Additionally, the range of hand actions required for assembly is relatively limited, simplifying the gesture recognition process.

Specifically, the actions involved in this project are divided into two primary types: picking up objects and tightening screws. The picking up action is further subdivided into two categories: "Catch Big" for larger components and "Catch Small" for smaller ones. Additionally, a "Done" gesture, indicating task completion, is also recognized. Therefore, the system is designed to detect a total of four key gestures.

To ensure accurate gesture recognition, two approaches were considered: traditional computer vision techniques and deep learning methods.

Traditional computer vision methods focus on extracting inherent features from the images, such as identifying key points or analyzing gradient information for differentiation. For instance, in detecting the "Done" gesture, the spatial relationship between the fingertips—specifically between the thumb and index finger—can be used. By examining their relative positions and the distribution of other fingers, the system can determine the gesture with reasonable accuracy.

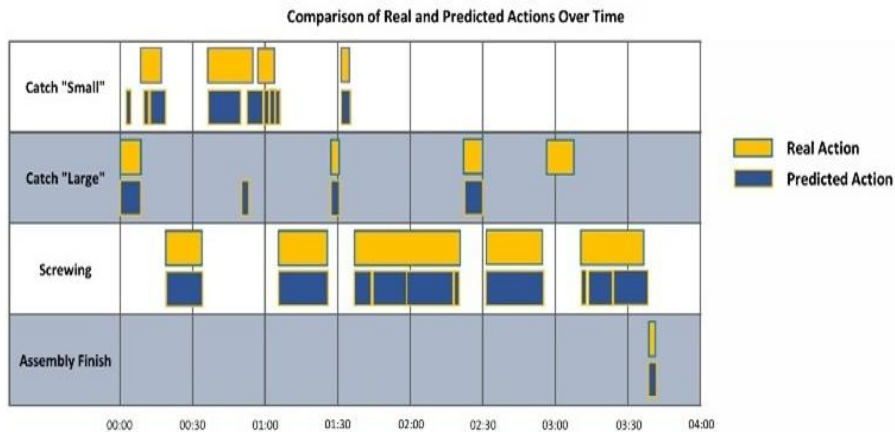

**Fig 6. Action prediction performance among 20 individual tests.** Blocks of different colors demonstrate precision of model intuitively.

However, for other gestures, traditional visual methods are insufficient. It is difficult and inefficient to define clear mathematical distinctions between various gestures, which also limits the flexibility and portability of the project. Therefore, a more robust machine learning-based solution is necessary for comprehensive gesture detection.

Considering the sequential nature of gesture data, this project implements an LSTM (Long Short-Term Memory) network for action recognition. LSTM models are well-suited for sequence data processing, effectively handling long-term dependencies and temporal patterns. The LSTM network was used to build the action recognition system, which demonstrated strong performance during training. Fig.6 illustrates the model's predictive performance throughout the complete assembly process.

To enhance data collection efficiency, the system captures frames at a rate of 30 frames per second using the camera. In each frame, the Mediapipe holistic module detects and stores key point data, resulting in 900 data points for each action over 30 frames. While this approach may introduce slight overfitting and limit accuracy to some extent, it significantly improves the system's time efficiency, enabling quick model training and testing.

### 3.6   Stage Detection and Verification

After implementing the deep learning models, we successfully obtained key data on object positions, orientations, depths, and action confidence levels. To ensure accurate stage detection and error handling during the manual assembly process, a Finite State Machine (FSM) was employed to structure the decision-making logic for verifying each assembly stage.

In actual operation, an assembly typically takes around 3 minutes and 30 seconds. Throughout this process, validation at each stage involves evaluating the positional relationships between components, including both plane coordinates and depth overlap, to avoid any misinterpretations of the assembly status. After completing a stage, the system checks whether the required components for the next stage have been correctly grasped by simultaneously verifying item attributes, positions, and action categories. If all criteria are met, the process moves forward to the screw hole detection stage. If any discrepancies are detected, the system provides visual feedback via interface pop-ups or notifications to guide the user in correcting the error.

As part of the system's design, the real-time image feed from the mobile camera is transmitted back to the local system via IP communication for processing. Screw hole statuses, captured through multithreaded methods, are sent back to the control system to facilitate joint validation. During the screw tightening process, the system continuously monitors the action. If the screw is not detected as fully tightened after the action, visual guidance is again used to correct the error. Once all screw holes for a given stage are verified, the assembly process moves on to the next stage.

To manage the detection, verification, and decision-making processes, we employed an FSM to govern the various states of the assembly. An FSM is a mathematical model used to represent systems with a finite number of states, providing a clear and concise way to model and control the stepwise transitions between different stages of the assembly.

In this project, the assembly process is divided into 21 sequential stages. It starts with the assembly of the actuator base, followed by screwing the electro component and verifying the screws for the arm and electro. Next, the actuator cover is assembled and screwed in, with subsequent verification of the screws. The platter and spindle are then assembled and secured, and their screws are verified. After confirming the inner components, the case cover and logic board are assembled, screwed, and inspected. Finally, the outer components are checked, leading to the final verification of the assembly and completion of the task.

The transitions between these states are based on key metrics, including the detected positional data, depth information, and action confidence scores. The system monitors the following key factors:

- **Det**: A nested array containing coordinates, depth, and confidence levels for each component.
- **Acf**: An array of confidence levels for each action.
- **Obj_angle**: An integer representing the specific angle of the actuator arm.

Whenever a state transition occurs, or if an error is detected, the system immediately provides feedback through the user interface and console. Error messages would guide the operator in resolving the issue, and detailed logs are recorded for further analysis and verification of the process. This FSM-based approach ensures a structured and reliable framework for real-time stage detection, error identification, and assembly verification, allowing for efficient and accurate monitoring of the entire process.

## 4  Experiments

In this section, we conduct a series of experiments to evaluate the performance and effectiveness of our proposed system in the context of manual assembly detection. The experiments focus on several key aspects, including object detection, action recognition, stage verification using a finite state machine (FSM), and overall system performance.

### 4.1  Dataset Evaluation

In this section, we evaluate the dataset used for training and testing the object detection and action recognition models. The dataset consists of images from various assembly scenarios, annotated for different components and actions.

To increase the diversity and robustness of the dataset, several data augmentation techniques were applied. These augmentations help simulate different lighting conditions, angles, and noise levels that may be encountered in real-world scenarios, improving the model's generalization ability.

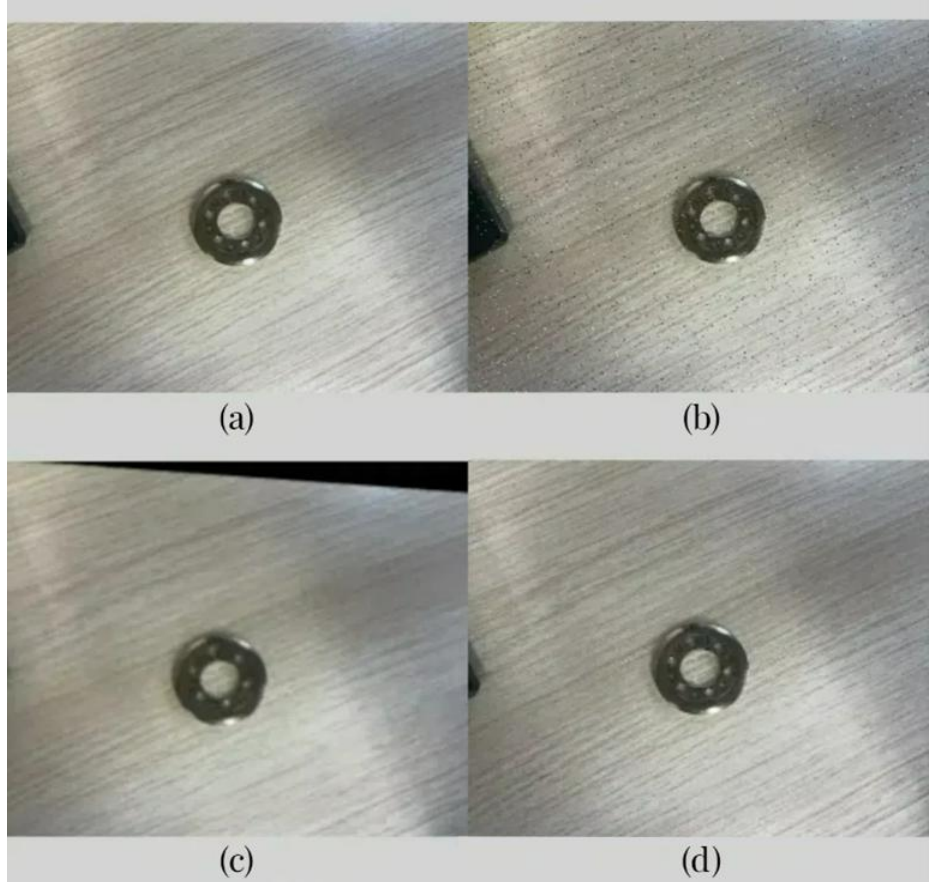

**Fig 7.** Dataset augmentation examples. From (a) to (d), the pictures represent original picture, noise-added picture, sheared picture, over saturated picture respectively.

Fig. 7 illustrates an example of the (a)original image and three of all the augmentation methods used in this project: (b)noise addition, (c)shearing, (d)saturation adjustment. Although other augmentation techniques, such as flipping, rotation, and scaling, were also utilized while they are not depicted in the figure. These methods collectively aim to increase the robustness of the model by introducing variability into the training data.

In addition to the visual transformations, Table. 1 presents a comparison between dataset before and after augmentation, showing the increased number of samples for each object class.

Especially, ArmElectro is underrepresented before augmentation, leading to substandard performance in anterior model evaluation. The data augmentation process not only increases the volume of training data but also ensures a balanced distribution across all object categories, which is crucial for maintaining the model's detection accuracy across different components.

**Table 1.** Dataset size comparison. The number of different classes are listed in two columns. Data size after augmentation increases around 40%.

| Classes | Num of Class | |
|---|---|---|
| | Original | Post-Augmentation |
| HDDCase | 528 | 737 |
| ActuatorArm | 198 | 363 |
| Platter | 165 | 317 |
| Screw | 108 | 258 |
| ActuatorBase | 114 | 240 |
| ActuatorCover | 112 | 239 |
| CaseCover | 112 | 239 |
| Spindle | 106 | 234 |
| LogiBoard | 108 | 233 |
| ArmElectro | **86** | 203 |

### 4.2 Object Detection Performance

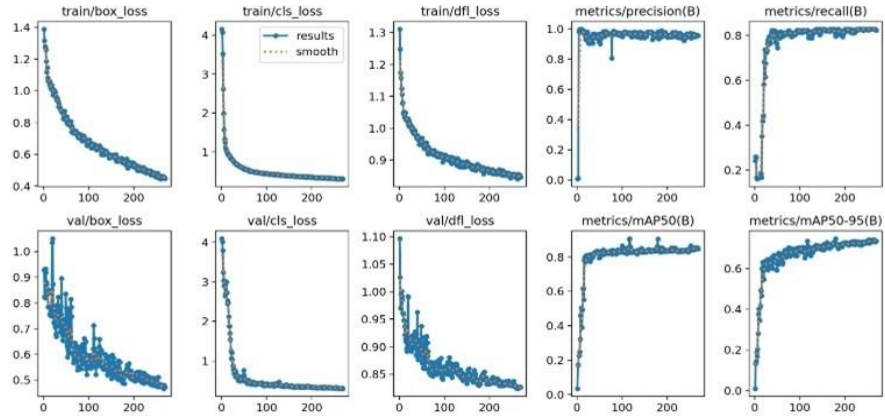

**Fig 8.** Object detection model metrics, showing training/validation loss curves and key performance metrics (precision, recall, mAP), highlighting model convergence and performance.

In the object detection task, we utilized the YOLOv5 model for detecting specific components in the assembly process. The model was trained and evaluated on a dataset containing various object types and orientations.

Throughout training, the model's performance was tracked by monitoring loss values (box loss, classification loss, and distribution focal loss) and evaluation metrics such as precision, recall, and mean average precision (mAP). As illustrated in Fig.8, the model demonstrates a steady reduction in both training and validation losses, with validation loss stabilizing after approximately 100 epochs. Precision and recall also show strong performance, achieving values around 0.9, while the mAP metrics (mAP50 and mAP50-95) indicate a robust model for detecting a range of objects with high accuracy.

### 4.3 Action Detection Model Evaluation

In this section, we evaluate the performance of the LSTM-based action detection model designed to classify hand gestures from key point sequences. The model processes 40 input nodes, each representing the (x, y, z, t) coordinates of a hand key point, and outputs one of four actions: Catch Big, Catch Small, Tightening, or Done.

**Network Structure.**
The proposed network is composed of three key components: LSTM layers to capture temporal dependencies, dense layers for feature aggregation, and a softmax output layer for classification. The input shape is (batch_size, 30, 40), representing a sequence of 30 frames with 40 key points (x, y, z, t) per frame.

The model starts with two LSTM layers: The first LSTM layer has 128 units, followed by a second LSTM layer with 256 units. These layers capture the temporal dynamics of the hand movements. A dropout rate of 0.2 is applied to each LSTM layer to prevent overfitting. Outputs of which are then flattened and passed through two fully connected (dense) layers: The first dense layer has 512 units with ReLU activation, followed by batch normalization and dropout (0.3). The second dense layer has 128 units, also with ReLU, batch normalization, and dropout (0.3).

The network outputs a 4-dimensional vector from a softmax layer, corresponding to the four possible hand gestures.

**Results and comparison.**

**Table 2.** Performance comparison of LSTM, 3D-CNN, and GRU across different sequence lengths (Short, Medium, Long) using Accuracy, Precision, Recall, and F1-Score.

| Metric     | Accuracy |        |        | F1-score |        |       |
|------------|----------|--------|--------|----------|--------|-------|
| Seq Length | Short    | Medium | Long   | Short    | Medium | Long  |
| Method     |          |        |        |          |        |       |
| 3D-CNN     | **0.9**  | 0.882  | 0.893  | 0.833    | 0.881  | 0.873 |
| GRU        | **0.9**  | 0.892  | 0.91   | **0.884**| 0.89   | 0.894 |
| Ours       | 0.874    | **0.922** | **0.0889** | 0.868 | **0.922** | **0.933** |

In Table 2, it compares the performance of three models (LSTM, 3D-CNN, and GRU) across different sequence lengths: Short (shorter than 20 frames), Medium (20 – 50 frames), and Long (more than 50 frames).

Across all models, we observe a general trend: longer sequences tend to result in better performance across accuracy, precision, recall, and F1-score. This improvement can be attributed to the fact that longer sequences provide more information about the hand gestures, allowing the models to capture richer temporal patterns. In contrast, shorter sequences offer less context, making it more challenging for the models to correctly identify the gestures, which is reflected in the generally lower scores for short sequences.

Among tests, LSTM consistently outperforms the other models across all sequence lengths, but the gap becomes more pronounced for medium and long sequences. The LSTM's ability to retain long-term dependencies in sequential data allows it to make more accurate predictions, especially in scenarios where actions unfold over multiple frames. Its high recall and F1-scores suggest that it not only identifies gestures correctly but also minimizes false negatives.

### 4.4 Angle Detection Model Evaluation

Using the detection and communication scheme, a YOLOv5-based distributed detection solution was implemented, with training on an RTX 2070 SUPER GPU taking about 1 hour to converge. The confusion matrix showed strong performance, achieving recognition accuracies of 99.07%, 89.13%, and 85.33% for three states. However, the model's accuracy drops significantly with major changes in lighting or background conditions.Assembly Stage Verification System Application

### 4.5 Assembly Stage Verification System Application

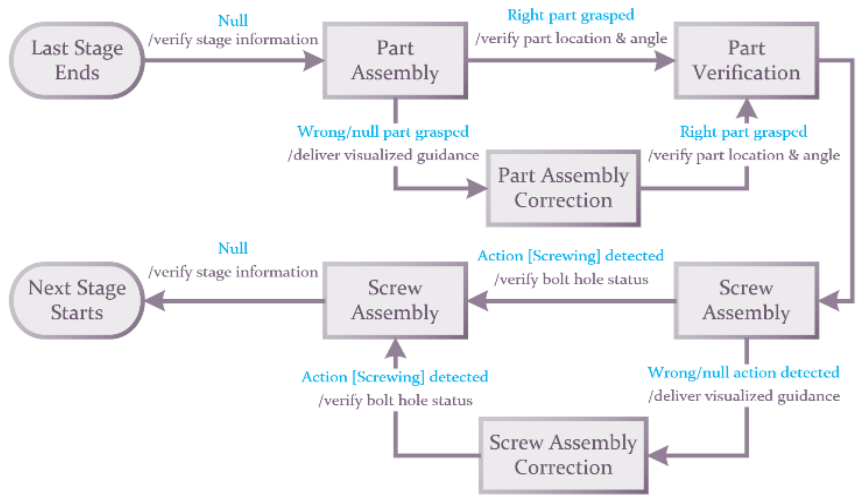

**Fig 9.** FSM-based assembly verification, ensuring accurate progression by detecting errors or null actions and providing visual guidance for corrections.

The performance of the assembly verification system is based on the Finite State Machine (FSM) approach, which enables precise monitoring and error correction at each critical step of the assembly process. As illustrated in Fig.9, the FSM-based system transitions between different states, ensuring that each stage in the assembly workflow is completed correctly before moving to the next.

The process begins once the previous stage has ended, where the system checks for null conditions and verifies the stage information. In the Part Assembly state, the system confirms that the correct part has been grasped and verifies its location and angle. If any error or null condition is detected (such as the wrong part being grasped), the system transitions to the Part Assembly Correction state, providing visualized guidance to the operator for error correction. Once the correct part is verified, the system moves forward to Part Verification, ensuring that all components are correctly positioned.

Similarly, during the Screw Assembly state, the system detects the action (such as screwing) and verifies the status of the bolt holes. If any incorrect or null action is detected, the system transitions to Screw Assembly Correction, delivering real-time feedback to the operator. Upon successful correction, the system resumes the normal workflow, proceeding to the next stage.

## 5 Conclusion

This article primarily introduces a visual solution that comprehensively applies various detection methods to detect and provide real-time feedback on errors during assembly stages. The key technologies and concepts involved include digital image processing, deep learning, filtering algorithms, stereo vision, finite state machines, Qt, and more.

Compared to other technical solutions in similar scenarios, this article possesses the following advantages:

- Different deep learning models are selected, and specific tuning parameters are applied based on the different characteristics of various objects. This ensures relatively timely and stable detection of each element, providing a solid data foundation for subsequent verification stages.
- The datasets used in various training stages of this system are collected under as diverse background conditions as possible and processed through data augmentation techniques. Combined with precise transition conditions defined in finite state machines, this system achieves relatively stable detection performance in different scenarios.
- Except for the inevitable information collection of various components in the assembly body, other parts of the system optimize the collection process through rich image processing techniques, reducing training requirements. Additionally, the distributed control strategy to some extent reduces the consumption of computing resources, further lowering the deployment requirements and facilitating application in different environments.

However, specifying the transition conditions for each stage through a state machine still somewhat limits the flexibility of the system. A better approach might be to integrate the data used during the detection process and use a Transformer structure to learn the correlations between different data sequence structures, thus defining different assembly stages and identifying errors.